\documentclass{article}

\usepackage{bm}
\usepackage{amsmath}

\usepackage{arxiv}

\usepackage[utf8]{inputenc} % allow utf-8 input
\usepackage[T1]{fontenc}    % use 8-bit T1 fonts
\usepackage{hyperref}       % hyperlinks
\usepackage{url}            % simple URL typesetting
\usepackage{booktabs}       % professional-quality tables
\usepackage{amsfonts}       % blackboard math symbols
\usepackage{nicefrac}       % compact symbols for 1/2, etc.
\usepackage{microtype}      % microtypography
\usepackage{lipsum}		% Can be removed after putting your text content
\usepackage{graphicx}
\usepackage{natbib}
\usepackage{doi}

\title{A 3D pocket-aware and evolutionary conserved interaction guided diffusion model for molecular optimization}

\date{} 					% Or removing it

\author{Anjie Qiao$^1$, Hao Zhang$^2$, Qianmu Yuan$^1$, Qirui Deng$^2$, Jingtian Su$^2$, Weifeng Huang$^2$, Huihao Zhou$^2$, \\ \textbf{Guo-Bo Li$^{3}$,  Zhen Wang$^{1}$,  Jinping Lei$^{2}$} \\
\\
\normalsize{$^{1}$School of Computer Science and Engineering, Sun Yat-sen University, 510006 Guangzhou, China}\\
\normalsize{$^{2}$School of Pharmaceutical Science, Sun Yat-sen University, 510006 Guangzhou, China}\\
\normalsize{$^{3}$West China School of Pharmacy, Sichuan University, 610041 Chengdu, China}\\
\normalsize{\{wangzh665@mail.sysu.edu.cn, leijp@mail.sysu.edu.cn.\}}
}

\hypersetup{
pdftitle={A template for the arxiv style},
pdfsubject={q-bio.NC, q-bio.QM},
pdfauthor={David S.~Hippocampus, Elias D.~Striatum},
pdfkeywords={First keyword, Second keyword, More},
}

\begin{document}
\maketitle

\begin{abstract}
	Generating molecules that bind to specific protein targets via diffusion models has shown good promise for structure-based drug design and molecule optimization. Especially, the diffusion models with binding interaction guidance enables molecule generation with high affinity through forming favorable interaction within protein pocket. However, the generated molecules may not form interactions with the highly conserved residues, which are important for protein functions and bioactivities of the ligands. Herein, we developed a new 3D target-aware diffusion model DiffDecip, which explicitly incorporates the protein-ligand binding interactions and evolutionary conservation information of protein residues into both diffusion and sampling process, for molecule optimization through scaffold decoration. The model performance revealed that DiffDecip outperforms baseline model DiffDec on molecule optimization towards higher affinity through forming more non-covalent interactions with highly conserved residues in the protein pocket. 
\end{abstract}

% keywords can be removed

\section{ Introduction}
\label{sec.intro}
Deep learning based generative models~\cite{1_wu2024tamgen,2_zhang2023learning,3_xie2025accelerating,4_chen2025deep,5_jiang2024pocketflow} have enabled fast exploration of vast chemical space and structure-based drug design without costly evaluation of a great many compounds compared with traditional process~\cite{6_zhang2025artificial}. In addition, generative models can directly generate novel drug-like molecules that might not be found within the medicinal chemists’ experiences~\cite{7_swanson2024generative}.

Recent developed 3D target-aware diffusion models~\cite{8_schneuing2024structure,9_igashov2024equivariant,10_huang2024dual,11_qian2024kgdiff,12_diffdec} that learn the joint distributions of all ligand atoms within the 3D protein pocket have been shown great promise for structure-based drug design and molecule optimization inside the protein pocket. For example, DiffSBDD and DiffLinker use diffusion models to generate molecules inside 3D protein pocket by iteratively denoising them from noise, and our recent developed DiffDec employed an E(3)-equivariant 3D conditional diffusion model for protein pocket-based molecular scaffold decoration. 

Although 3D target-aware diffusion models have been applied to both \textit{de novo} molecular generation and molecular optimization, their core mathematical formulation remains largely the same across these tasks~\cite{28_targetdiff,14_huang2024protein,29_decompdiff,12_diffdec,18_luo20213d,36_diffbp}. In \textit{de novo} design, generation is typically conditioned on a  target protein pocket, while molecular optimization extends this by also conditioning on known ligand fragments. This suggests that, to generate molecules with more realistic drug-like properties, effective conformations, and authentic binding modes, relying solely on these modeling approaches may not be sufficient.

Therefore, to improve control over the generation process, some tasks integrate expert knowledge or additional awareness-based guidance~\cite{30_diffint,31_zhung20243d,32_wu2024guided,33_lee2024ncidiff}. Especially, the 3D target-aware diffusion models with explicit binding interaction guidance, such as InterDiff~\cite{13_wu2024guided}, IPDIFF~\cite{14_huang2024protein}, and IRDIFF~\cite{15_huang2024interaction}, have been most recently proposed for generating molecules that form specific interactions with amino acid residues inside the protein pocket. However, a major limitation of these interaction guided diffusion models is that the generated molecules may not form interactions with the highly conserved residues, which are important for protein functions and bioactivities of the ligands~\cite{16_sumida2024improving}.

To overcome this challenge, we herein developed a new 3D target-aware diffusion model \textbf{DiffDecip} with explicit protein-ligand binding interaction and evolutionary conservation guidance for molecular optimization through scaffold decoration. 
Our model introduces two key mechanisms:~\textbf{(1) Conservation-Aware Condition}, which leverages residue conservation scores to encourage the generation of R-groups that form more interactions with highly conserved residues; and~\textbf{(2) Interaction-Prior Guidance}, which uses a pretrained interaction-prior network to guide the generation process based on learned protein-ligand interaction patterns. 

We build upon DiffDec as the base model and integrate these two mechanisms.
Specifically, an interaction-prior IPNet~\cite{14_huang2024protein} network was pretrained to capture the protein-ligand interactions, and the evolutionary conservation scores of protein residues were served as additional conditions. The model performance revealed that DiffDecip outperforms DiffDec on generating molecules that form more interactions with the highly conserved residues for improving the binding affinity.
\begin{figure}[t]
    \centering
    \includegraphics[width=0.9\linewidth]{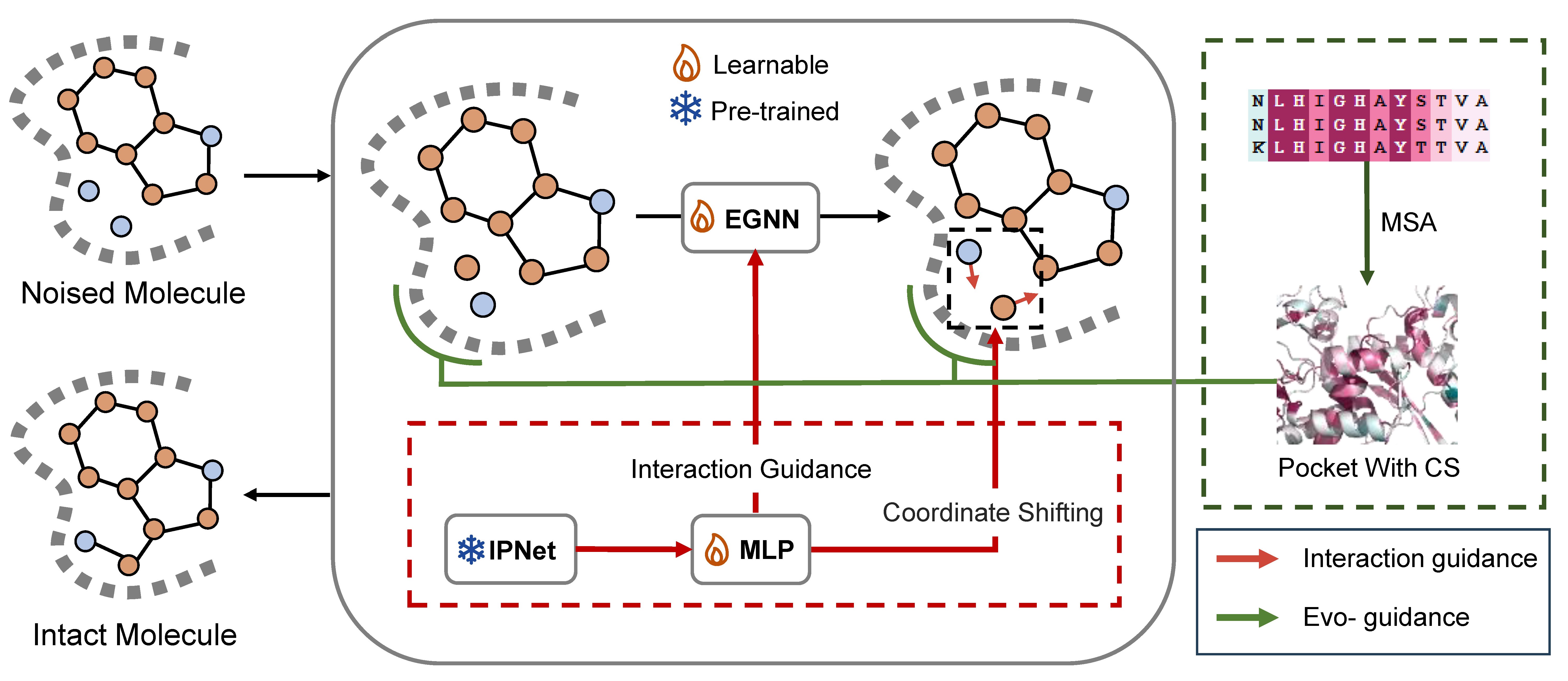}
    \caption{Model Overview.}
    \label{fig:framework}
\end{figure}
\section{Materials and Methods}
\label{sec.methods}
\subsection{Dataset Construction}
In this study, we focus on single R-group decoration-based molecule generation, where the goal is to generate an R-group that attaches to a predefined molecular scaffold. To achieve this, we conduct our experiments using the CrossDocked data set~\cite{17_crossdocked} and adopt the same preparation and reaction-based slicing method as Diffdec~\cite{12_diffdec}. Specifically, the CrossDocked dataset contains 22.5 million docked binding complexes. At first, we selected high-quality docking poses (RMSD < 1$\mathring{A}$) and ensured protein diversity (sequence identity < 30\%). 

Next, we extracted the protein substructure within a 10$\mathring{A}$ radius of the binding ligand. Following the splitting settings proposed by Luo et al.~\cite{18_luo20213d}, the original training and test sets consisted of 100,000 and 100 protein-ligand pairs, respectively. Each ligand in the original training and test sets was split into a scaffold and an R-group using the method from LibINVENT~\cite{19_fialkova2021libinvent}, which enhances validity and synthetic accessibility by applying 37 customized reaction-based rules. This process yielded a final training set of 76,370 tuples and a test set of 49 tuples, with each tuple comprising a protein pocket, a scaffold, and an R-group.

\subsection{Preliminary}
Given a protein pocket $\mathcal{P} = \{(x_i^\mathcal{P}, v_i^\mathcal{P})\}_{i=1}^{N_P}$ and a ligand molecule $\mathcal{M} = (\mathcal{S}, \mathcal{R})$, in this work, a ligand molecule $\mathcal{M}$ can be represented as a scaffold $\mathcal{S} = \{(x_i^\mathcal{S}, v_i^\mathcal{S})\}_{i=1}^{N_S}$ and an R-group $\mathcal{R} = \{(x_i^\mathcal{R}, v_i^\mathcal{R})\}_{i=1}^{N_R}$. 
Where $N_P$ (resp. $N_S$, $N_R$) refers to the number of atoms of the pocket $\mathcal{P}$ (resp. the scaffold $\mathcal{S}$, the R-group $\mathcal{R}$). $x \in \mathbb{R}^3$ and $v \in \mathbb{R}^K$ denote the 3D position coordinate and type of the atom, respectively, where $K$ denotes the number of atom types with $v$ being one-hot vector. Throughout this work, matrices are represented by uppercase boldface letters. For a given matrix $\bm{X}$, $x_i$ denotes the vector in its $i$-th row.

For brevity, we denote the pocket as 
\[
\bm{P} = [\bm{X}^\mathcal{P}, \bm{V}^\mathcal{P}],
\quad \text{where } \bm{X}^\mathcal{P} \in \mathbb{R}^{N_P \times 3}, \; \bm{V}^\mathcal{P} \in \mathbb{R}^{N_P \times K}.
\]
Similarly, the scaffold and R-group are represented as 
\[
\bm{S} = [\bm{X}^\mathcal{S}, \bm{V}^\mathcal{S}], \quad \bm{R} = [\bm{X}^\mathcal{R}, \bm{V}^\mathcal{R}],~\text{respectively}.
\]
 The decoration task can be formulated as modeling the conditional distribution $p\left(R \middle| P, S\right),$
where, for simplicity, our notation does not distinguish the random variables from their realizations (i.e., specific matrices). 

We proposed a conservation- and interaction-guided diffusion model to generate an R-group that is expected to bind to the conserved regions of the given pockets while enhancing interaction within the complex. Our framework extends the diffusion model with two key mechanisms:~\textbf{(1) Conservation-aware conditioning pocket}, which encourages the R-group to grow toward conserved region of the binding pocket, and~\textbf{(2) Interaction-prior guidance}, which promotes the formation of stronger and more diverse interactions between the decorated ligand and the pocket.

\subsection{Conservation-Aware Condition}
For each protein pocket $\mathbf{P} = [\mathbf{X}^\mathcal{P}, \mathbf{V}^\mathcal{P}]$, we first retrieve its full amino acid sequence using its UniProt ID~\cite{20_uniprot}. We then compute sequence conservation by performing multiple sequence alignment (MSA) against the UniRef30\_2023\_02 database~\cite{21_uniref} through HHblits~\cite{22_hhblits}. This process yields a conservation score matrix $\mathbf{C}^\mathcal{P}$ for the pocket $\mathbf{P}$, where $\mathbf{C}^\mathcal{P} \in \mathbb{R}^{N_P \times 1}$ and each score $c_i \in [0, 1]$, in which a larger value indicates the $i$-th amino acid residue with higher conservation. The conservation scores serve as additional features for the pocket representation, resulting in an augmented pocket descriptor:
\[
\mathbf{P}_{\mathbf{CA}} = [\mathbf{X}^\mathcal{P}, \mathbf{V}^\mathcal{P}, \mathbf{C}^\mathcal{P}].
\]

\subsection{Interaction-Prior Guidance}
We leverage IPNet~\cite{14_huang2024protein}to offer interaction-based prior to our diffusion model. Specifically, IPNet, denoted as $\psi_{\text{IPNet}}(\cdot)$, is a pretrained interaction-prior network built upon SE(3)-equivariant neural networks~\cite{23_SE3} and cross-attention layers~\cite{24_cross-attention} to predict the binding affinity of protein-ligand complexes.

Given a protein $\bm{P}$ and a ligand molecule $\bm{M}$, IPNet encodes their interactions into protein and ligand representations, capturing the complex interplay between protein binding sites and ligand molecules:
\[
F^\mathcal{P}, F^\mathcal{M} = \psi_{\text{IPNet}}(\bm{P}, \bm{M})
\]
This representation is then utilized to predict the binding affinity of the protein-ligand complex. In our work, as the focus is on molecular optimization, we decompose each ligand molecule into a scaffold and an R-group. The representation obtained from IPNet is further repurposed as an interaction prior to guide the diffusion process. Specifically, given a pocket $\bm{P}$, scaffold $\bm{S}$, and R-group $\bm{R}$, we extract their interaction-prior representations as:
\[
F^\mathcal{P}, F^\mathcal{S}, F^\mathcal{R} = \psi_{\text{IPNet}}(\bm{P}, \bm{S}, \bm{R})
\]
Next, we incorporate the \textbf{Conservation-Aware} and \textbf{Interaction-Prior} mechanisms to guide both the forward diffusion and reverse denoising stages for molecule generation.

\subsection{Forward Diffusion Process}
Recent studies often adopt fixed form noise schedule for diffusion-based generative models~\cite{26_diffusion,38_lipman2022flow,39_song2020score}. In our work, we employ a variance-preserving cosine schedule version~\cite{25_schedule1,26_diffusion}, which defines a set of $\beta_t\ (t=0,\ldots,T)$ for each timestep $t$. Rather than directly applying this schedule, we redefine the noise process following the Signal-to-Noise Ratio (SNR)~\cite{27_SNR}:
\[
\gamma_t = \log\left(1 - \beta_t\right) - \log \beta_t,
\]
\[
\alpha_t = \sqrt{\mathrm{sigmoid}\left(-\gamma_t\right)},
\]
\[
\sigma_t = \sqrt{\mathrm{sigmoid}\left(\gamma_t\right)}.
\]
In the forward process, we model the atomic coordinates and types of $\mathbf{R}$ as continuous random variables and introduce noise iteratively from a Gaussian distribution $\mathcal{N}(0, I)$ at each time step $t$. The scaffold $\mathbf{S}$ and pocket $\mathbf{P}$ are treated as fixed contextual information, remaining unchanged throughout both the forward and reverse processes:
\[
q\left(R_t \middle| R_0, S, P\right) = \mathcal{N}\left(R_t; \alpha_t R_0, \sigma_t^2 I\right).
\]

Moreover, we can extract the interactive representations of R-group $F_0^\mathcal{R}$ from the pretrained IPNet $\psi_{\text{IPNet}}\left(R_0, S, P\right)$. We then introduce a learnable neural network $\psi_\theta(\cdot)$ for guiding the R-group atom coordinates with an interaction-based shifting:
\[
S_t^\mathcal{R} = k_t \cdot \psi_\theta\left(F^\mathcal{R}, t\right),
\]
where $\psi_\theta(\cdot)$ is a MLP neural network, $S_t^\mathcal{R} \in \mathbb{R}^{N_R \times 3}$ is the cumulative mean shift in step $t$, and $k_t$ is a pre-defined preserving scaling coefficient:
\[
k_t = \sqrt{\prod_{s=1}^{t}\left(1 - \beta_s\right)} \cdot \left(1 - \sqrt{\prod_{s=1}^{t}\left(1 - \beta_s\right)}\right).
\]

We incorporate this mean shift into the forward diffusion process as follows:
\[
q\left(R_t \middle| R_0, S, P, F_0^\mathcal{R}\right) = \mathcal{N}\left(R_t; \alpha_t R_0 + S_t^\mathcal{R}, \sigma_t^2 I\right),
\]
\[
q\left(R_t \middle| R_{t-1}, S, P, F_0^\mathcal{R}\right) = \mathcal{N}\left(R_t; \alpha_{t|t-1}\left(R_{t-1} - S_{t-1}^\mathcal{R}\right) + S_t^\mathcal{R}, \sigma_{t|t-1}^2 I\right),
\]
where
\[
\alpha_{t|t-1} = \frac{\alpha_t}{\alpha_{t-1}},
\]
\[
\sigma_{t|t-1}^2 = \sigma_t^2 - \alpha_{t|t-1}^2 \sigma_{t-1}^2.
\]

\subsection{Reverse denoising process}
In the reverse process, our objective is to learn to reverse the forward noise injection. The corresponding posterior can be expressed as:
\[
q\left(R_{t-1} \middle| R_t, S, P\right) = \mathcal{N}\left(R_{t-1}; \widetilde{\mu}\left(R_t, R_0, t\right), \widetilde{\sigma}\left(t\right) I\right),
\]
where
\[
\widetilde{\mu}\left(R_t, R_0, t\right) = \frac{\alpha_{t|t-1} \sigma_{t-1}^2}{\sigma_t^2} R_t + \frac{\alpha_{t-1} \sigma_{t|t-1}^2}{\sigma_t^2} R_0,
\]
\[
\widetilde{\sigma}\left(t\right) = \sigma_{t|t-1}^2 \sigma_{t-1}^2 / \sigma_t^2.
\]

However, since the ground-truth fragment $R_0$ is inaccessible at timestep $t$, we train a denoiser $\varphi_\theta(\cdot)$ to approximate it. Specifically, we use the denoiser to predict the Gaussian noise:
\[
\hat{\epsilon}_t = \varphi_\theta(R_t, S, P, t),
\]
and the training objective is to minimize the mean squared error between true noise $\epsilon$ and the predicted noise $\hat{\epsilon}_t$:
\[
\mathcal{L} = ||\epsilon - \hat{\epsilon}_t||^2.
\]

From the predicted noise, we obtain an estimate of the ground-truth:
\[
\hat{R}_0 = \left(1/\alpha_t\right) R_t - \left(\sigma_t/\alpha_t\right) \hat{\epsilon}_t.
\]

Then, we replace $P$ with the Conservation-augmented pocket $P_{CA}$ when predicting the noise:
\[
\hat{R}_0 = \left(1/\alpha_t\right) R_t - \left(\sigma_t/\alpha_t\right) \varphi_\theta(R_t, S, P_{CA}, t).
\]

Since $R_0$ is not directly available, we use the estimated $\hat{R}^{t+1}_0$ from the previous time step $t+1$ to compute the interactive representation:
\[
F_{t+1}^\mathcal{R}, F_{t+1}^\mathcal{P}, F_{t+1}^\mathcal{S} = \psi_{IPNet}(P, S, \hat{R}_0^{t+1}),
\]
which are then used to calculate the mean shift for the R-group:
\[
S_t^\mathcal{R} = k_t \cdot \psi_\theta(F_{t+1}^\mathcal{R}, t).
\]

Similarly, the estimate $\hat{R}_0^t$ obtained at timestep $t$ is used to compute $S_{t-1}^\mathcal{R}$ at next timestep $t-1$. At the beginning timestep $t = T$, the initial interactive representation $F_{T+1}^\mathcal{R}$ is `None`.

To maximize the exploitation of the protein-ligand interaction prior encoded in the pre-trained IPNet, we further integrate the interactive representation $F_{t+1}^\mathcal{P}$, $F_{t+1}^\mathcal{S}$, $F_{t+1}^\mathcal{R}$ into the denoiser prediction:
\[
\hat{R}_0 = \left(1/\alpha_t\right) R_t - \left(\sigma_t/\alpha_t\right) \varphi_\theta(R_t, S, P_{CA}, F_{t+1}^\mathcal{P}, F_{t+1}^\mathcal{S}, F_{t+1}^\mathcal{R}, t).
\]

Therefore, the reverse transition kernel is then formulated as:
\[
p\left(R_{t-1} \middle| R_t, S, P, F_{t+1}^\mathcal{R}, F_{t+1}^\mathcal{P}, F_{t+1}^\mathcal{S} \right) = \mathcal{N}\left(R_{t-1}; \widetilde{\mu}\left(R_t, \hat{R}_0, F_{t+1}^\mathcal{R}, F_{t+1}^\mathcal{P}, F_{t+1}^\mathcal{S}, t\right), \widetilde{\sigma}(t) I\right),
\]
where
\[
\widetilde{\mu}\left(R_t, R_0, F_{t+1}^\mathcal{R}, F_{t+1}^\mathcal{P}, F_{t+1}^\mathcal{S}, t\right) = \frac{\alpha_{t|t-1} \sigma_{t-1}^2}{\sigma_t^2}(R_t - S_t^\mathcal{R}) + \frac{\alpha_{t-1} \sigma_{t|t-1}^2}{\sigma_t^2} \hat{R}_0 + S_{t-1}^\mathcal{R}.
\]

In this way, by integrating both the Conservation-Aware and Interaction-Prior mechanisms, our diffusion model leverages evolutionary information and interaction cues to guide the reverse process to recover the underlying molecular configuration.

\section{Experiments}
\label{sec.results}

\subsection{Baselines}
In this work, we adopt DiffDec as the base model and equip it with the two proposed mechanisms, forming our enhanced model, DiffDecip. Accordingly, our evaluation focuses on comparing the performance of DiffDecip with DiffDec. DiffDec is an E(3)-equivariant 3D-conditional diffusion model for scaffold decoration that incorporates 3D protein pocket constraints without domain knowledge guidance and only employs diffusion on atom types and coordinates. This approach has been successfully employed for the structure-based scaffold decoration.

\subsection{Metrics}
We evaluate our models on the test set. For each protein pocket, we generate 100 candidate molecules and report the average values across a range of evaluation metrics:
\begin{itemize}
    \item \textbf{Validity}: The percentage of generated molecules that preserve the original scaffold and can be successfully parsed by RDKit~\cite{37_rdkit}.
    \item \textbf{Uniqueness}: The proportion of unique molecules among the generated molecules.
    \item \textbf{Vina Score}: An estimate of the binding affinity (in kcal/mol) between the generated molecules and the target protein. Molecular docking is  performed using QVina~\cite{34_QVina} to calculate the Vina scores.
    \item \textbf{High Affinity}: The percentage of test cases in which the generated molecules achieve Vina scores greater than to that of the reference compound.
\end{itemize}

Moreover, we evaluate the average frequency of interactions between the generated R-groups and protein pockets across 49 test set pockets, with interactions computed using Schrödinger tools~\cite{35_schrodinger2024}. Additionally, we identify a subset of test pockets containing highly conserved residues (conservation scores > 0.4) and, for each case, calculate the average number of interactions specifically between the R-groups and these conserved residues.

\subsection{Results}
DiffDecip (Fig.~\ref{fig:framework}) incorporates the guidance of protein-ligand interactions and evoluationary conservation information of amino acids in protein pocket into both diffusion and sampling process. Specifically, the conservation scores of the residues serve as additional features for the protein pocket representation, and an interaction-prior IPNet~\cite{14_huang2024protein} built upon SE(3)-equivariant neural networks and cross-attention layers is pretrained to capture the protein-ligand interactions through supervision of binding affinity signals. In the diffusion process, Gaussian noises are gradually added to the atomic coordinates and types of R-group through a variance-preserving cosine schedule~\cite{27_SNR}, and the protein-ligand interactions of R-group from pretrained IPNet are formulated as fixed conditions. In the sampling process, both the conservation scores of residues and protein-ligand interactions are integrated for adapting the positions of generated R-group to enhance non-bonded interactions with highly conserved residues.

\begin{figure}[t]
    \centering
    \includegraphics[width=1\linewidth]{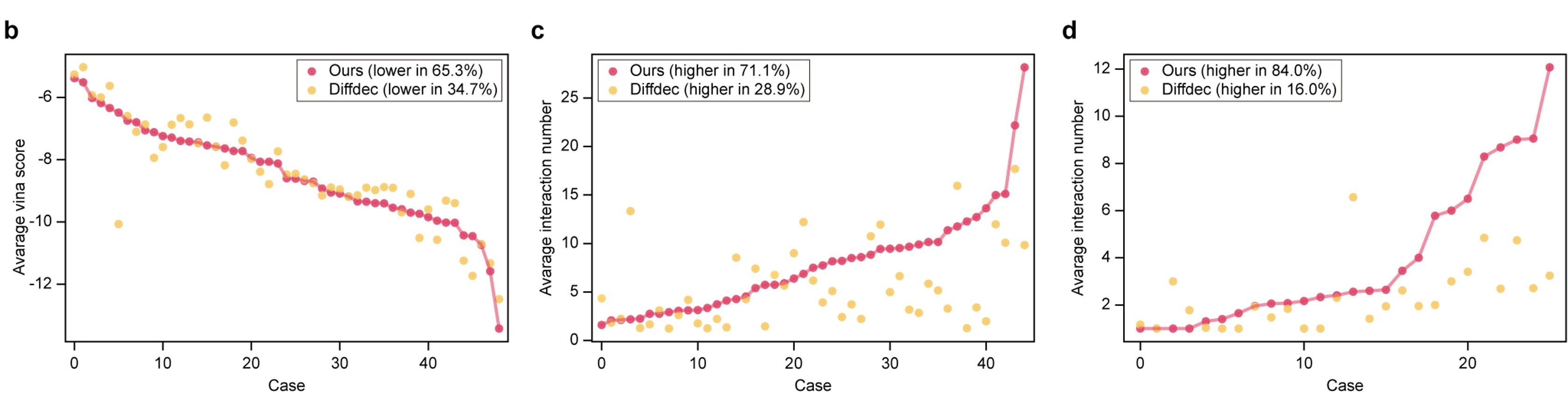}
    \caption{Performance of DiffDecip and comparison with DiffDec.}
    \label{fig:result}
\end{figure}

We evaluated and compared the performance of DiffDecip with our previously developed DiffDec model. As shown in Table~\ref{table1}, DiffDecip outperforms DiffDec on binding-related metrics including Vina score and High Affinity. Herein, Vina score is the most important binding-related metric as it directly evaluates the binding affinity of generated 3D molecules without optimizing the conformation. DiffDecip shows higher average of Vina score than DiffDec (-8.19 vs -8.11). When comparing binding affinities of the generated molecules with the reference ligands, 44.6\% of the molecules generated by DiffDecip exhibits higher binding affinity, while 42.9\% molecules generated by DiffDec exhibit higher binding affinity (Table~\ref{table1}).

The binding performance (Fig.~\ref{fig:result}b) on each of the 49 protein targets in test set showed that DiffDecip achieves the best binding affinity in 65.3\% protein targets, while DiffDec achieves best affinity in only 34.7\% targets. This is probably because the molecules generated by DiffDecip form more non-bonded interactions with amino acid residues in the protein pocket. As shown in Fig.~\ref{fig:result}c, molecules generated by DiffDecip form more interactions with the residues than those generated by DiffDec in 71.1\% protein targets of test set. Thus, the pretrained interaction-prior IPNet in DiffDecip successfully drive the molecule generation towards forming more non-bonded interactions with target protein for improving the binding affinity.

\begin{table}[h]
\centering
\caption{Model performance of DiffDecip compared with DiffDec.}
\begin{tabular}{c|cccc}
\bottomrule
          & Validity($\uparrow$) & Uniqueness($\uparrow$) & Vina Score($\downarrow$) & High Affinity($\uparrow$) \\ \bottomrule
DiffDec   & \textbf{91.8}        & 53.3          & -8.11         & 42.9             \\
DiffDecip & 90.1        & \textbf{67.3}          & \textbf{-8.19}         & \textbf{44.6}             \\ \bottomrule
\end{tabular}

\label{table1}
\end{table}

Notably, molecules generated by DiffDecip form more interactions with the highly conserved residues than those generated by DiffDec in 84\% protein targets, which is larger than the value of 71.1\% evaluated on all interacting residues (Fig.~\ref{fig:result}d). Thus, the improve of binding affinity for molecules generated by DiffDecip are greaterly due to the increased interactions with highly conserved residues. This indicates the joint guidance of protein-ligand interactions and evolutionary conservation information of amino acids in pocket successfully steers molecule generation towards higher affinity through forming more non-bonded interactions with highly conserved residues in the protein pocket.

\clearpage
\bibliographystyle{unsrtnat}
\bibliography{references}  %%% Uncomment this line and comment out the ``thebibliography'' section below to use the external .bib file (using bibtex) .





% \end{thebibliography}

\end{document}